\documentclass[conf]{new-aiaa}
\usepackage[utf8]{inputenc}
\usepackage{algpseudocode}
\usepackage{graphicx}
\usepackage{amsmath}
\usepackage[version=4]{mhchem}
\usepackage{siunitx}
\usepackage{longtable,tabularx}
\usepackage{xcolor} 

\usepackage{comment}
\usepackage{float}
\usepackage{caption}
\usepackage{subcaption}
\usepackage{multicol}
\usepackage{todonotes}
\usepackage{algpseudocode}
\usepackage{algorithm2e}
\RestyleAlgo{ruled}

\title{Scalable Planning and Learning Framework Development for Swarm-to-Swarm Engagement Problems}

\author{Umut Demir\footnote{Authors contributed equally to this work}, A. Sadik Satir*, Gulay Goktas Sever*, Cansu Yikilmaz* and N. Kemal Ure}
\affil{Istanbul Technical University, Artificial Intelligence and Data Science Application and Research Center, Istanbul, Turkey.}

\begin{document}

\maketitle

\begin{abstract}
Development of guidance, navigation and control frameworks/algorithms for swarms attracted significant attention in recent years. That being said, algorithms for planning swarm allocations/trajectories for engaging with enemy swarms is largely an understudied problem. Although small-scale scenarios can be addressed with tools from differential game theory, existing approaches fail to scale for large-scale multi-agent pursuit evasion (PE) scenarios. In this work, we propose a reinforcement learning (RL) based framework to decompose to large-scale swarm engagement problems into a number of independent multi-agent pursuit-evasion games. We simulate a variety of multi-agent PE scenarios, where finite time capture is guaranteed under certain conditions. The calculated PE statistics are provided as a reward signal to the high level allocation layer, which uses an RL algorithm to allocate controlled swarm units to eliminate enemy swarm units with maximum efficiency. We verify our approach in large-scale swarm-to-swarm engagement simulations.

%In certain conditions, guaranteed capture in finite time is provided by each multi-agent pursuit evasion games. We use t

% \textit{details regarding the low level framework may be mentioned here (PE game formulation, different strategies) - cansu}

% We evaluate our approach with simulation results to present superior scalability compared to existing approaches.

% \textit{no simulation is done to reflect the superior scalability of the proposed method yet 
% , may mention the effect of low level algorithms on the high level framework (improvement of the RL algorithm - cansu)
% }

\end{abstract}

% \section{Nomenclature}

% {\renewcommand\arraystretch{1.0}
% \noindent\begin{longtable*}{@{}l @{\quad=\quad} l@{}}
% $A$  & amplitude of oscillation \\
% $a$ &    cylinder diameter \\
% $C_p$& pressure coefficient \\
% $Cx$ & force coefficient in the \textit{x} direction \\
% $Cy$ & force coefficient in the \textit{y} direction \\
% c   & chord \\
% d$t$ & time step \\
% $Fx$ & $X$ component of the resultant pressure force acting on the vehicle \\
% $Fy$ & $Y$ component of the resultant pressure force acting on the vehicle \\
% $f, g$   & generic functions \\
% $h$  & height \\
% $i$  & time index during navigation \\
% $j$  & waypoint index \\
% $K$  & trailing-edge (TE) nondimensional angular deflection rate
% \end{longtable*}}

\section{Introduction}
\vspace{0.3cm}

Swarm-to-Swarm engagement problem consists of computing optimal assignments/allocations and  strategies of controlled swarm units against an adversarial swarm\cite{uzun2020probabilistic}. Various high level swarm guidance algorithms are developed in recent years, using both deterministic \cite{eren2017velocity,zhao2011density} and probabilistic methods \cite{demir2015decentralized,uzun2020probabilistic,uzun2020decentralized}. Although these studies present scalable results for the high level guidance problem, they either ignore the low-level dynamics of the engagements or do not involve any engagements at all (such as only focusing on waypoint guidance tasks). We argue that ignoring the low level dynamics, such as the evasion strategy of the adversarial swarm units or the pursuit strategy of the controlled swarm units, can lead to sub-optimal assignments, which in turn leads to inefficient guidance strategies for the controlled swarm.
    
On the other hand, swarm-to-swarm engagement problem can be entirely formulated in the framework of
multi-agent pursuit-evasion (PE) differential game\cite{isaacs1999differential}. Although differential game theory provides a promising formal approach to address pursuit evasion problems\cite{bacsar1999stochastic}, it is computationally infeasible to scale these approaches beyond a few pursuers and evaders.
% , which makes it incompatible with swarm-to-swarm engagement problems that might involve tens of agents. The aforementioned swarm pursuit-evasion scenario has unique challenges. 
The aforementioned pursuit-evasion scenarios has unique challenges. 
In addition to computing optimal pursuer policy, it also required to compute optimal assignment of large number of pursuers to evaders, which is a combinatorial decision problem over a large number of possible assignments. Due to its combinatorial nature, large scale swarm-to-swarm engagement problems are infeasible to attack with the conventional PE methods.

The main goal of this paper is to utilize recent advances in reinforcement learning (RL) and differential game theory to
improve the state of the art performance in swarm to swarm engagement problem. We show that this objective can be achieved by decomposing the problem into two layers; i) a high level decision making layer for allocating controlled swarm units using RL, ii) a low level guidance layer for solving allocated PR problems with tools from differential geometry and providing a reward signal to higher layer for making more informed decisions on engagement allocations.

% \newpage
% \subsection{Previous Work on Pursuit-Evasion Methods}
\subsection{Previous Work}
Pursuit-evasion games define a certain class of problems that involve one or more pursuers attempting to intercept one or more evaders. In general, game is over when pursuers reach a certain region around the evader. 
There are many kind of multi-agent pursuit evasion games \cite{chung2011search}: convex or non-convex environment \cite{alexander2006pursuit}, continuous or discrete time \cite{adler2003randomized}, bounded or unbounded, holonomic or nonholonomic \cite{kothari2017cooperative}, constraints \cite{eklund2011switched}, speed, and sensing. 
Pure-distance pursuit strategy which is directly minimizes the instantaneous distance to the evader. This strategy provides an optimal pursuit strategy for certain zero-sum differential games in unbounded environments \cite{isaacs1999differential}.
Although, in continuous case optimal solution to minimize time-to-capture can be obtained by using Hamilton–Jacobi–Isaacs (HJI) partial differential equations, its computational complexity creates a limitation on the number of agents in the game.  
The Voronoi partition approach is an effective method for holonomic systems to compute safe-reachable areas when all the players have identical speed constraints \cite{gavrilova2008generalized}.
For holonomic systems, multiple pursuers- single evader with identical speed constraints in a convex domain was studied using safe reachable area minimization policy for a 2D environment with a guaranteed capture \cite{guaranteed_decentralized,pan2012pursuit,zhou2016cooperative}.
Pierson et al. \cite{inter_rogue_robots} extended this results to N dimensional space and presented decentralized version of this algorithm.
For non-holonomic systems, Kothari et al \cite{kothari2017cooperative}. proposed  a computationally
efficient algorithm to extend area minimization strategy to non-holonomic systems. 
% In our framework we employed pure-distance pursuit policy  and area minimization policy as presented in \cite{inter_rogue_robots}.\kXX{Bu son cumleyi buraya koymayalim, alttaki PE subsectioninda bahsedelim.}

% \subsection{***Previously on Reinforcement Learning***}
% Reinforcement learning is a way of intelligently learning actions to maximize a cumulative reward. Although, the RL problem is fundamentally the same,  the environment poses a variety of distinct problems. The existence of successful learning initiatives in various environments such ah AlphaGo and AlphaZero shows that suitable Reinforcement learning algorithms can solve challenging environments. Although the subjects in single or multi agent problems are studied well, RL for swarm intelligence is an open problem.

Reinforcement learning is a branch of machine learning that aims to learn control strategies for sequential decision making problems by relying on interactions with the system, without the explicit knowledge of the underlying dynamics. In recent years, there has been significant improvement in RL algorithms for solving highly complex decision making problems~\cite{silver2016mastering,silver2018general}. That being said, extension of these algorithms to control of large-scale multi-agent problems\cite{vinyals2019grandmaster,foerster2018counterfactual,sunehag2017value,rashid2018qmix} remain as an open challenge due to scalability limitations.

%  Although the RL problems are fundamentally the same; the environments pose various problems. The existence of
% successful learning initiatives in various environments such ah AlphaGo \cite{silver2016mastering} and AlphaZero \cite{silver2018general} shows that appropriate Reinforcement learning algorithms can address these kinds of challenges.
% Multi-agent environments where multiple learning agents must work together to optimize a single cumulative reward are one of the challenging problems in this field.

\subsection{Contribution}
In this work, we propose an RL based framework to overcome limitations of current techniques on swarm to swarm engagement problem. Our framework is composed of two layers; at high level optimal assignments and decomposition of sub-games are handled by an RL algorithm, in low level independent small scale  multi-agent pursuit evasion games are solved by differential game theoretical approaches.\\
Solution of each PE game provide several performance metrics, such as average, maximum and minimum capture time information to compute a reward signal, which is used by RL algorithm to compute engagement allocations. This information flow inherently connects pursuit-evasion game performance with overall performance of the engagement algorithm. Thus the high level algorithm can ensure efficient allocations, without explicitly modeling the low level dynamics, since the reward signal represents a compressed version of the complex interaction between pursuers and evaders.

\begin{figure}[H]
    \centering
    \includegraphics[width=\textwidth]{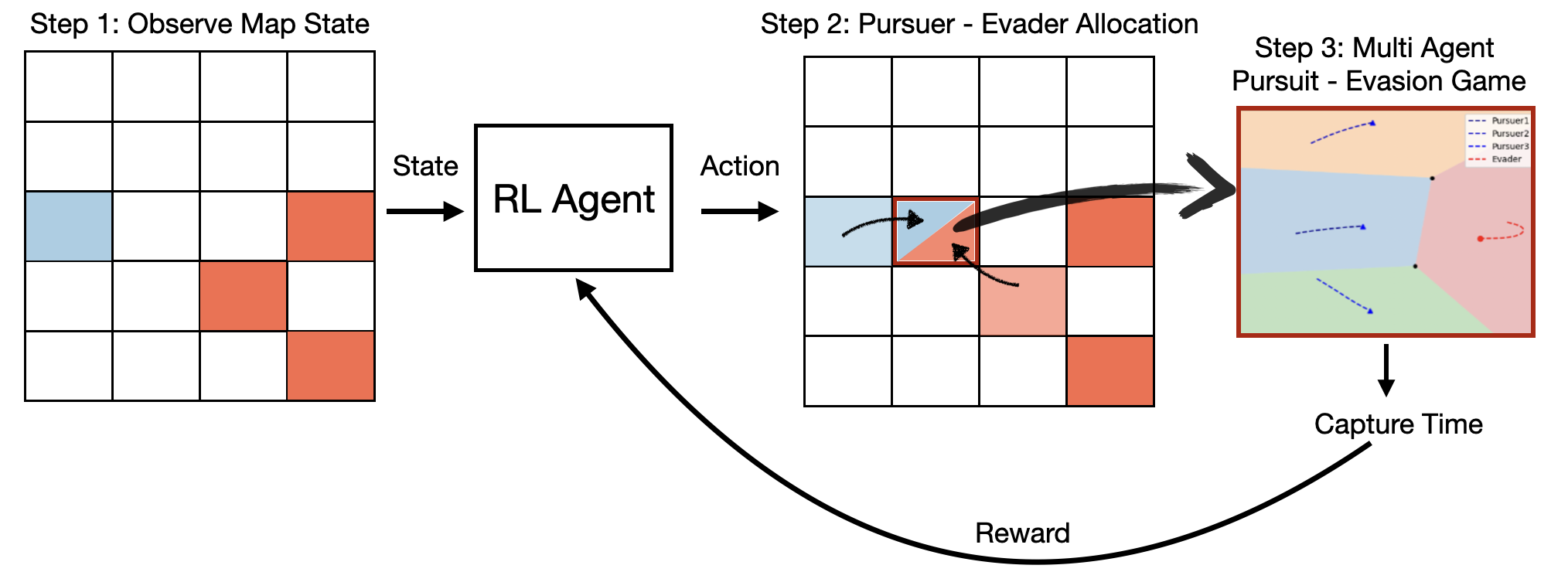}
    \caption{The proposed swarm-to-swarm learning and planning framework. The controlled swarm density is represented by blue grids, whereas the adversarial swarm occupies the red grids. The high level RL algorithm decides allocation of engagements between units, whereas the low level algorithm solves the independent multi-agent PE games at each grid. Two layers are connected to each other through the reward signal, which carries performance metrics from the solution of the low level PE problem.}
    \label{fig:rl_arc}
\end{figure}

High-level overview of the architecture given in Fig.\ref{fig:rl_arc}.

This paper is organized as follows; the Section 2 gives a multi-agent pursuit evasion game strategies, the Section 3 provides the details of the high level allocation of agents with Reinforcement Learning, finally Section 4 gives simulation results for various cases.

\section{Low Level Multi-Agent Pursuit Evasion Game}
\hspace{0.1cm}

In our framework we employed pure-distance pursuit policy  and area minimization policy as presented in \cite{inter_rogue_robots}.\\
Consider a multi-player pursuit-evasion game with a group of $n_p$ pursuer and $n_e$ evader agents in a bounded, convex environment $ Q \subset \mathbb{R}^{2}$ with points given as $q$. We assume all agents have the following equations of motion \cite{inter_rogue_robots} 

\begin{eqnarray}
\begin{split}
\dot{x}_e^i &= u_e^i, \quad \Vert u_e(t) \Vert \leq v_{\max },\\
\ \dot{x}_p^j &= u_p^j, \quad \Vert up(t) \Vert \leq v_{\max }, 
\label{eq:1}
\end{split}
\end{eqnarray}

where ${x}_e^i$ denote the positions of evaders for $i \in \{1,...,n_e\}$, ${x}_p^j$ denote the positions of pursuers for $j \in \{1,...,n_p\}$. ${u}_e$ and ${u}_p$ represent the control inputs of evaders and pursuers, respectively, and both subject to identical maximum velocity $v_{\max }$. We assume that $v_{\max} = 1$ for the rest of the paper. 

Define the capture condition of the evader $i$ at $x_e^i(t_c)$ for the pursuers as,

\begin{eqnarray} 
\min _{j \in n_p} \Vert x_e^i(t_c) - x_p^j(t_c) \Vert < r_c, \label{eq:2}
\end{eqnarray}

where $r_c > 0$ is the capture radius.

\subsection{Pursuit Strategies}
\subsubsection{Area-Minimization Policy}
In this section, we briefly describe the area-minimization policy for pursuers to capture an evader by decreasing its safe reachable area which is denoted as $A_e$, in consistent with the existing literature. $A_e$ is defined with points that the evader can reach before any other agent. Throughout the section, we adapt the notation used in \cite{inter_rogue_robots}.

Safe reachable area of the evader reduces to its Voronoi cell under the assumption of pursuers and evaders both having equal speeds. Voronoi tessellation is defined in \ref{eq:voronoi_tessellation}

\begin{eqnarray} 
V_i = \lbrace q \in Q | \Vert q - x^i \Vert \leq \Vert q - x^j \Vert, \forall j \ne i, \;\;i,j \leq n \rbrace. 
\label{eq:voronoi_tessellation}
\end{eqnarray}

The area of the Voronoi cell $V_e$ containing the evader is calculated as follows,

\begin{eqnarray} 
A_e = \int _{V_e} dq,
\label{eq:voronoi_area}
\end{eqnarray}

The dynamics of $A_e$ are given by,

\begin{eqnarray}
\dot{A}_e = \frac{\partial A_e}{\partial x_e} \dot{x}_e + \sum _{j=1}^{n_p} \frac{\partial A_e}{\partial x_p^j} \dot{x}_p^j. 
\label{eq:cell_dynamics}
\end{eqnarray}

According to the given dynamics \eqref{eq:cell_dynamics}, pursuers cooperatively try to minimize the time derivative of $A_e$ while the evader try to maximize it as an opposite act of action. The joint objective for pursuers may reduce to minimization of each agent's individual contribution $\frac{\partial A_e}{\partial x_p^j} \dot{x}_p^j$. To lead pursuers in the direction of the fastest decrease of the evader's area, we define the control input for the pursuer $j$ as,

\begin{eqnarray} 
u_p^j = - \frac{\frac{\partial A_e}{\partial x_p^j}}{\Vert \frac{\partial A_e}{\partial x_p^j} \Vert }. 
\label{eq:u_area_min}
\end{eqnarray}

The gradient $\frac{\partial A_e}{\partial x_p^j}$ is calculated as,

\begin{eqnarray}
\frac{\partial A_e}{\partial x_p^j} = \frac{L_j}{\Vert x_p^j - x_e \Vert }\left(x_p^j - C_{b_j} \right).
\label{eq:area_gradient}
\end{eqnarray}

where $L_j$ is the area, $C_{b_j}$ is the centroid of the boundary. 
Details regarding the derivation of the gradient can be found in \cite{inter_rogue_robots}.

Substituting \eqref{eq:area_gradient} into \eqref{eq:u_area_min}, control policy of the pursuer reduces to

\begin{eqnarray}
u_p^j = \frac{\left(C_{b_j} - x_p^j \right)}{\Vert C_{b_j} - x_p^j \Vert }. 
\label{eq:u_purs_min}
\end{eqnarray}

\subsubsection{Pure Distance Pursuit}
Pure-distance pursuit policy implies that a pursuer moves directly towards its nearest evader ${e}^\kappa$ of which $\kappa$ is determined from the expression\cite{inter_rogue_robots}

\begin{eqnarray}
\kappa = \text{arg min}_{i \in n_e}{\Vert x_{e}^\kappa - x_p^j \Vert }.
\label{eq:9}
\end{eqnarray}

Control policy is calculated as,

\begin{eqnarray} 
u_p^j = \frac{\left(x_{e}^\kappa - x_p^j \right)}{\Vert x_{e}^\kappa - x_p^j \Vert }. \label{eq:10}
\end{eqnarray}

\subsection{Evasion Strategies}
\subsubsection{Constant Area Policy}
Under the area-minimization strategy, the area $A_e$ is always non-increasing until the capture condition is met for any admissible control strategy\cite{guaranteed_decentralized}. Only control policy that the evader can apply to keep the area constant is given as,

\begin{eqnarray}
u_e^* = \frac{\left(C_{b} - x_e \right)}{\Vert C_{b} - x_e \Vert },
\label{eq:11}
\end{eqnarray}

where $C_b$ indicates the middle point of the shared Voronoi boundary between pursuer $x_p$ and evader $x_e$.

\subsubsection{Move-to-centroid Policy}
In accordance with the \cite{inter_rogue_robots}
we choose to use move-to-centroid as one of the policies of the evader.

\begin{eqnarray} 
u_e^i = \frac{\left(C_{V_i} - x_e^i \right)}{\Vert C_{V_i} - x_e^i \Vert },
\label{eq:12}
\end{eqnarray}

where $V_i$ is the Voronoi region of the evader and $C_{V_i}$ is the centroid. Under this policy, evader can avoid not only one but multiple pursuers by moving towards the centroid and getting away from its cell's edges.

%%%%%%%%%%%% HIGH LEVEL FRAMEWORK  %%%%%%%%%%%%%%%%%%%%%%
\section{High Level Pursuit Evasion Allocation with Reinforcement Learning}
\hspace{0.1cm}

We consider a Markov Decision Process~(MDP) \cite{sutton2018reinforcement} with discrete time steps where a single agent observes full map state and generates continuous actions until termination conditions are satisfied. The MDP consists of 5 parameters $\langle S,A,T,R,\gamma \rangle$, where  $S$ is the state space of a $N \times N$ grid map, A is the action space, $T(S_{t+1}|S_t,A_t)$ is the transition distribution and $\gamma \in [0,1)$ is the discount factor and $R$ is the reward obtained from the transition. The agent actions are determined with a policy function $\pi (a \mid s; \theta)$ parameterized by $\theta$ that outputs continuous actions in the interval of $ a_{low} \leq a \leq a_{high} $.
The main objective of the problem is to maximize the discounted cumulative reward,
\begin{equation}
    \label{eq:exp}
    \displaystyle \mathop{\mathbb{E}}_{\pi, \gamma} \left[\sum_{i=0}^{T} \gamma^{i} R_i \mid S_0 = s \right]
\end{equation} 
The state space, $S$, of the MDP consists of density distribution over the $NxN$ grid map. 

\begin{equation}
 S =  
\begin{pmatrix}
 s_1      &s_2      & ...    &s_N   \\ 
 s_{N+1}  &s_{N+2}  & ...    &...  \\ 
 ...      &...      & ...    &...   \\
 s_{N^2 - N + 1}    & ...    & ...    &s_{N^2} 
\end{pmatrix}_{N \times N}   \text{ where }  0 \leq p \leq 1 \text{  } s \in S \text{ and } \sum_{p \in S} p = 1
\end{equation}

In the vector form;

\begin{equation}
 S^T =  
\begin{pmatrix}
 s_1              &s_2     & ...&s_N &
 s_{N+1} & s_{N+2}  & ... & s_{N^2} 
\end{pmatrix}_{1 \times N^2}   \text{ where }  0 \leq s \leq 1 \text{  } s \in S \text{ and } \sum_{s \in S} s = 1
\end{equation}

We consider a state transition matrix, $T$, which gives probabilities of passing from one density distribution to the next.
\begin{equation}
 T = 
\begin{pmatrix}
 t_{1,1} & t_{1,2} & ... &  t_{1,N^2} \\ 
 t_{2,N^2+1} & t_{2,N^2+2}  & ... & ...\\ 
 ... & ... & ...  & ... \\
  ...  & .... & ... & t_{N^2,N^2} \\ 
\end{pmatrix}_{N^2 \times N^2}   \text{where }  0 \leq t \leq 1 \text{  } t \in T \text{ and } \sum_{i = 1 }^{N} T(i,j) = 1 \text{ for j=1,...,}N^2
\end{equation}

\begin{equation}
    S'_{k+1} = T_k*S_k \text{ where k is time step, } k = 0,1,..,k_{max}
\end{equation}

It is necessary to find transition matrices that maximize the Equation \ref{eq:exp}. The number of elements of the transition matrix increases as $N^4$ and it becomes too complex even for small sized problems. To reduce size of the action space, the nature of the problem is used and the transition from one grid to another is limited with its neighbors. Further, diagonal and self transitions are allowed. Therefore, corners have 4, edge, non-corner grids have 6 and inner grids have 9 possible actions. The number of actions, $N_{actions}$, generated by policy, $\pi (a\mid s)$ can be calculated as:
\begin{equation}
    N_{actions} = 4*4 + 6*4(N-2) + 9*(N-2)^2
\end{equation}

The action space of the MDP, $A$, represented as follows;

\begin{equation}
    A = 
    \begin{pmatrix}
     a_1 & a_2 & a_3 & ... & a_{N_{actions}}
    \end{pmatrix}_{ 1 \times N_{actions}}
\end{equation}
Finally, infeasible transitions in the transition matrix are set to zero. The transition matrix in terms of actions can be written as follows;
\begin{equation}
    T =
    \begin{pmatrix}
 a_1 &  a_2 & 0 & ... & 0 & a_{n+1} &  a_{n+2} & 0 & ... &  0 \\ 
 a_{n+3} & a_{n+4}  & 0 & .  & . & . & .  & . &. &.\\ 
 . & . & .  & .  & . & .  & . & . &. &.\\
 0 & . &. &. &. &. & 0 & a_{N_{actions} - 1} & a_{N_{actions}}\\ 
\end{pmatrix}_{N^2 \times N^2}
\end{equation}

The reward function is combination of both number of intruder units and average capture time.
\begin{equation}
    R_k = -C_{distribution}\sum_{n=1}^{N} s_{n, enemy}  + C_{capture}*time_{capture}
\end{equation}
where $C_{distribution}$ and $C_{capture}$ are the coefficients that controls the weights of individual rewards. $time_{capture}$ is normalized values from Table \ref{table:pp_tima_capt} and $S_{enemy}$ is the state of the density distribution over the map. 

% To compute a policy for the swarm allocation MDP, we use the Twin Delayed DDPG algorithm~(see cite{fujimoto2018addressing} \cite{TD3}), which is presented in Alg. \ref{alg:TD3}.
\begin{algorithm}
\caption{Twin Delayed DDPG \cite{fujimoto2018addressing} \cite{TD3}} \label{alg:TD3} 
Input: Initial Policy Parameters $\theta$, Q-function parameters $\phi_1, \phi_2$, empty replay buffer $D$ \\
Set target parameters to initial parameters $\theta_{target} \gets \theta$, $\phi_{target, 1} \gets \phi_1 $, $\phi_{target,2} \gets \phi_2$ \\
\textbf{Repeat} \\
Select an action: $a = \mu_{\theta(s)} + \epsilon $, where $s$ is state, $ \epsilon \sim \mathcal{N} $ and $ a \in [a_{low}, a_{high}]$ \\
Execute action $a$ and obtain next state, $s'$ \\
Store $(s,a,r,s',d)$ replay buffer, $D$ \\
If $s'$ is terminal, reset the environment \\
\If{it's time to update}{
        \For{j in range} {Randomly sample a batch of transitions, $B=(s,a,r,s',d)$ from $D$ \\
        Compute target actions $a'(s') = \mu_{\theta(s')} + clip( \epsilon, -c, c)  $ \\
        Compute targets \\ $y(r, s', d) = r + \gamma(1-d)\underset{i=1,2}{\text{min}} Q_{\phi_{target,i}}(s', a'(s'))$ \\
        \If{$j \text{ mod } \text{policy delay} == 0$}{
        Update Q-function with gradient descent using \\ $\nabla_{\theta_i}\frac{1}{\mid B \mid} \sum_{(s,a,r,s',d) \in B}  ( Q_{\phi_i}(s,a) - y(r,s',d))^2 $ \\
        Update target networks with \\
        $\phi_{target,i} \gets \rho \phi_{target,i} + (1-\rho)\phi_{i} $ \\
        $\theta_{target,i} \gets \rho \theta_{target,i} + (1-\rho)\theta_{i} $ } } }
\textbf{Until}{Converge}
\end{algorithm}

% \newpage
%\section{Problem Formulation}
%\input{sections/problem_formulation}
\section{Results}
\vspace{0.3cm}
In this section, we discuss low level pursuit evasion games and  high level allocation results  along and provide details of the simulation environment.

\subsection{Low-Level Pursuit-Evasion Strategies}
\label{sec:pusuitresults}
In this section, simulation studies are conducted by comparing
generalized area minimization policy with pure-distance pursuit for different number of pursuers. 

The simulations are performed in a 10 x 10 square, with all players moving at a maximum speed of 1, capture radius of 0.25, and a fixed time step of 0.01 in Python.
% \todo{simulasyonların hangi ortamda yapıldığını yazalım mı}
For the pursuer strategy, we apply area-minimization and pure-distance pursuit policies, whereas the evader either moves towards the target point or try to keep their safe reachable area constant. Note that there are no assumptions regarding the evader's strategies. 50 Monte Carlo simulations are completed to illustrate the performance of each strategy and  analyze the variation of capture time with different number of pursuers.
% #######################%
\subsubsection{Pure Distance Pursuit Strategy}
\begin{figure}[H]
     \centering
     \begin{subfigure}[b]{0.3\textwidth}
         \centering
         \includegraphics[width=\textwidth]{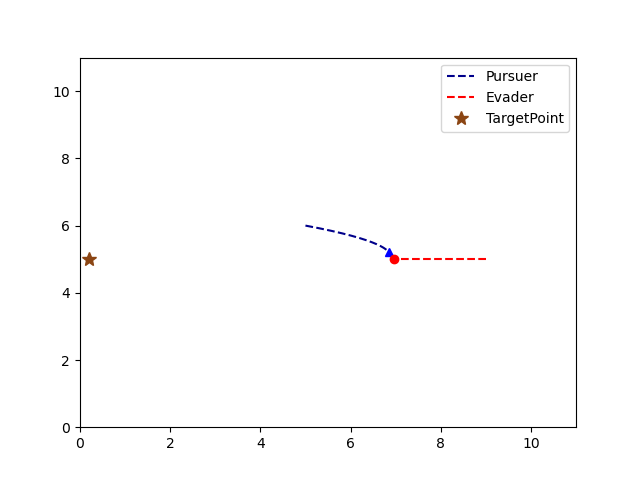}
         \caption{Single pursuer and evader}
        % \caption{Tekli kaçan ve kovalayan}
         \label{fig:1x1_dc}
     \end{subfigure}
     \hfill
     \begin{subfigure}[b]{0.3\textwidth}
         \centering
         \includegraphics[width=\textwidth]{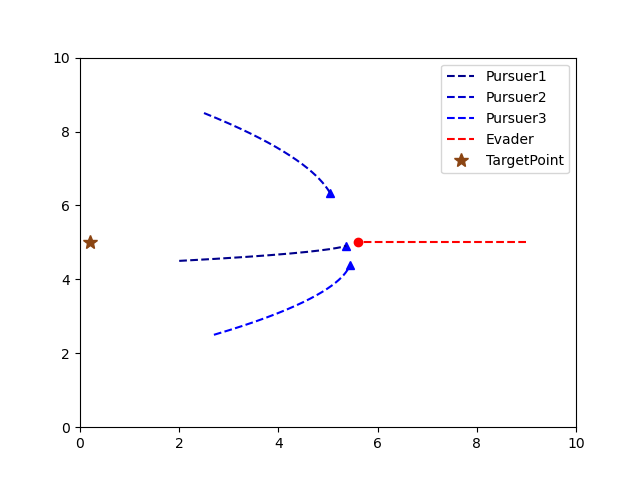}
         \caption{3 pursuers and one evader}
        % \caption{3 kovalayan,tek kaçan}
         \label{fig:3x1_dc}
     \end{subfigure}
     \hfill
     \begin{subfigure}[b]{0.3\textwidth}
         \centering
         \includegraphics[width=\textwidth]{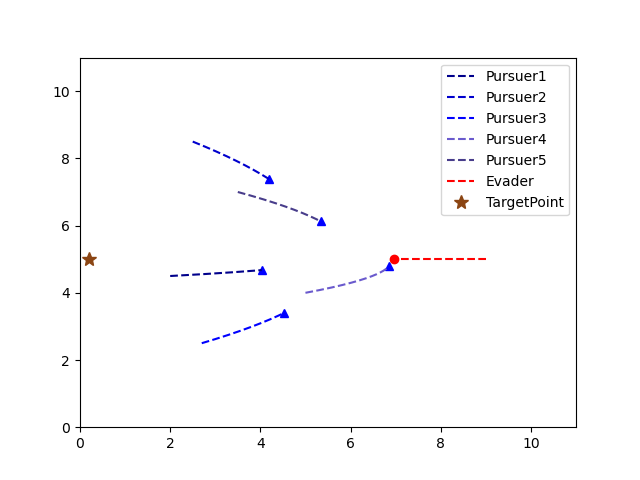}
         \caption{5 pursuers and evader}
        % \caption{5 kovalayan,tek kaçan}
         \label{fig:5x1_dc}
     \end{subfigure}
        \caption{Trajectories for pursuers(blue triangle) and evader (red circle) under the pure-distance pursuit strategy. Evader moves towards to target point pursuers directly towards to evader  }
        \label{fig:all_pp}
\end{figure}

Under pure-distance pursuit strategy each pursuers directly minimizes the instantaneous distance to the evader. It is an optimal strategy for unbounded environments. In Fig.\ref{fig:all_pp}, evader minimizes distance with constant target point and each pursuer move towards to current position of the evader. Both use pure-distance pursuit strategy.

The results for 1-3-5 pursuers to single evader shown in Fig.\ref{fig:all_pp}.

\begin{table}[H]
\caption{Capture time comparison results obtained by MC simulations}
\begin{center}
\begin{tabular}{|c|c|}
    \hline
    \multicolumn{2}{c}{Pure-distance pursuit strategy}\\
    \hline
    Case & Average time(sec)\\
    \hline
    1 pursuer & 3.373\\
    \hline
    3 pursuers & 2.856\\
    \hline
    5 pursuers & 2.0\\
\hline
\end{tabular}
\end{center}
\label{table:pp_tima_capt}
\end{table}

For each MC simulations, the initial conditions of each pursuer is drawn from a uniform distribution $x\in [3,5]$ and  $y\in [0,10]$; and for the evaders $x\in [7,9]$ and  $y\in [0,10]$.The tabulated results are shown in Table\ref{table:pp_tima_capt}.

%%%%%%%%%%%%%%%%%%%%
\subsubsection{Area Minimization Strategy}
In this section, fundamental strategy is same for all of the pursuers, which is area minimization through Voronoi decomposition; however, there is a distinction between 1-1, 3-1 and 5-1 cases in terms of evader's strategies. In the single pursuer case, evader apply the constant area policy to keep its Voronoi cell constant while in multiple pursuer cases, evader is directed to centroid of its safe reachable area. Algorithm \ref{alg:area_min} gives a high level overview of the our strategy.

\begin{algorithm}[H]
\caption{Global Single Evader Policy} \label{alg:voronoi} 
\While{$(\min _{j \in P_k} \Vert x_p^j - x_e^\kappa \Vert > r_c)$}{
Calculate Voronoi tessellation with all agents \\
Determine pursuers which are the Voronoi neighbours,  ${N}_e$ \\
\uIf{$x_p^j \in  {N}_e$}{
    Compute control input \ref{eq:u_area_min} \;
  }
  \Else{
    Compute control input \ref{eq:10} \;
  }}
\label{alg:area_min}
\end{algorithm}

% \begin{algorithm}[H]
% \caption{Global Tekli Kaçan Stratejisi} \label{alg:TD3} 
% \While{$(\min _{j \in P_k} \Vert x_p^j - x_e^\kappa \Vert > r_c)$}{
% Tüm ajanların Voronoi ayrıştırmasını yap \\
% Voronoi komşuları olan kovalayanalrı belirle,  {N}_e \\
% \uIf{$x_p^j \in  {N}_e$}{
%     Kontrol girdisi= \ref{eq:u_area_min} \;
%   }
%   \Else{
%     Kontrol girdisi=  \ref{eq:10} \;
%   }}
% \ENDFOR{}
% \ENDIF{}
% \ENDELSE{}
% \label{alg:area_min}
% \end{algorithm}

% Simulations are initialized with the assignment of random coordinates to each pursuer and evader inside the map. 
According to positions of players, the map area is decomposed into Voronoi regions which are then used to calculate the control policies for both opponents. The simulation process continues iteratively until the capture condition is met.

\begin{figure}[H]
     \centering
     \begin{subfigure}[b]{0.3\textwidth}
         \centering
         \includegraphics[width=\textwidth]{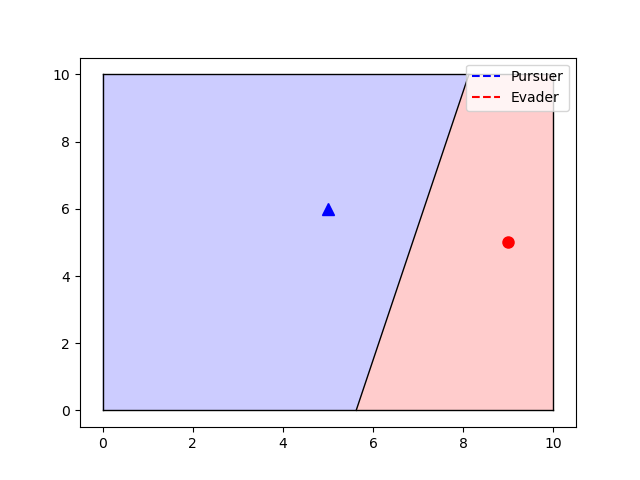}
         \caption{$t=0$s}
         \label{fig:1x1_dc_1}
     \end{subfigure}
     \hfill
     \begin{subfigure}[b]{0.3\textwidth}
         \centering
         \includegraphics[width=\textwidth]{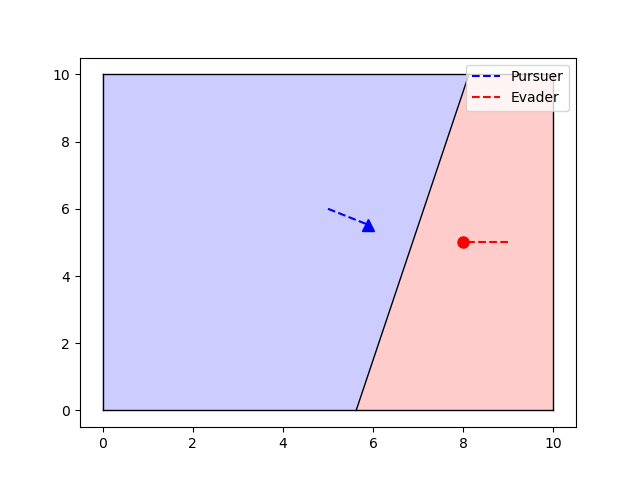}
         \caption{$t=1$s}
         \label{fig:3x1_dc_1}
     \end{subfigure}
     \hfill
     \begin{subfigure}[b]{0.3\textwidth}
         \centering
         \includegraphics[width=\textwidth]{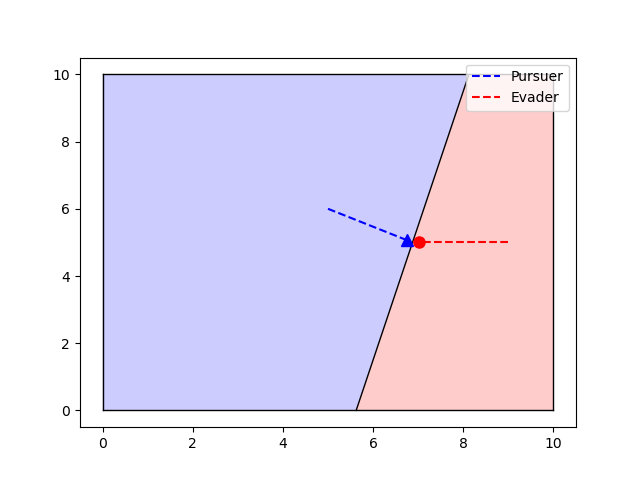}
         \caption{$t=2$s}
         \label{fig:5x1_dc_1}
     \end{subfigure}
        \caption{Trajectories for 1 pursuer(blue triangle) and 1 evader (red circle) under the area minimization strategy. Each shaded area represents Voronoi cell of the related point. }
        \label{fig:all_1x1_voronoi}
\end{figure}

\begin{figure}[H]
     \centering
     \begin{subfigure}[b]{0.3\textwidth}
         \centering
         \includegraphics[width=\textwidth]{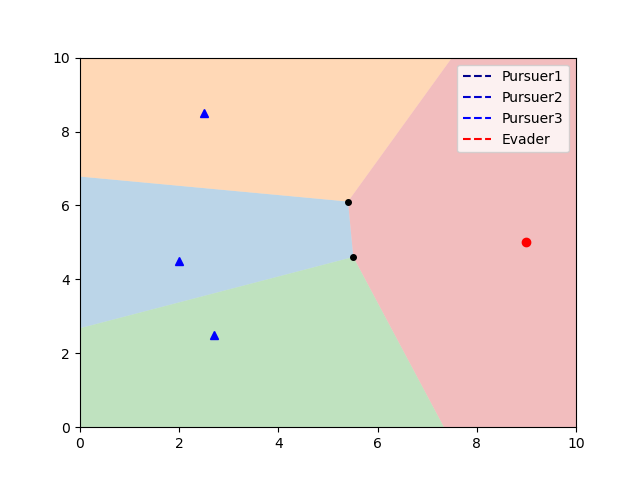}
         \caption{$t=0$s}
         \label{fig:1x1_dc_2}
     \end{subfigure}
     \hfill
     \begin{subfigure}[b]{0.3\textwidth}
         \centering
         \includegraphics[width=\textwidth]{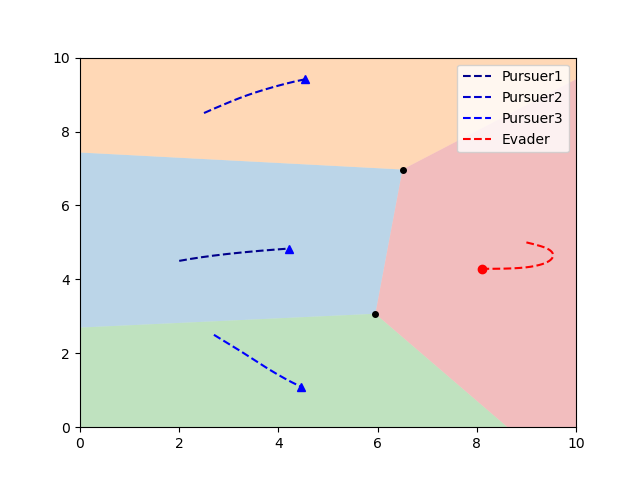}
         \caption{$t=2.24$s}
         \label{fig:3x1_dc_2}
     \end{subfigure}
     \hfill
     \begin{subfigure}[b]{0.3\textwidth}
         \centering
         \includegraphics[width=\textwidth]{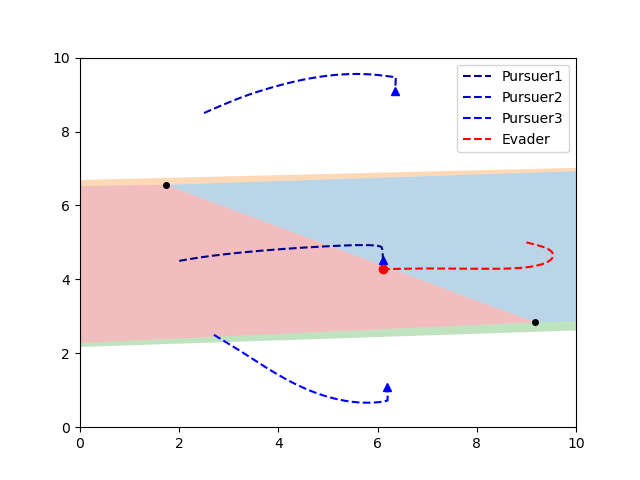}
         \caption{$t=4.47$s}
         \label{fig:5x1_dc_2}
     \end{subfigure}
        \caption{Trajectories for 3 pursuers(blue triangle) and 1 evader (red circle) under the area minimization strategy. Each shaded area represents Voronoi cell of the related point. }
        \label{fig:all_3x1_voronoi}
\end{figure}

\begin{figure}[H]
     \centering
     \begin{subfigure}[b]{0.3\textwidth}
         \centering
         \includegraphics[width=\textwidth]{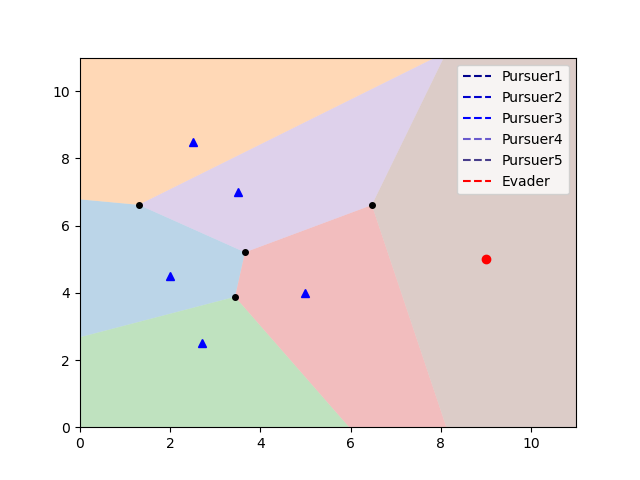}
         \caption{$t=0$s}
         \label{fig:1x1_dc_3}
     \end{subfigure}
     \hfill
     \begin{subfigure}[b]{0.3\textwidth}
         \centering
         \includegraphics[width=\textwidth]{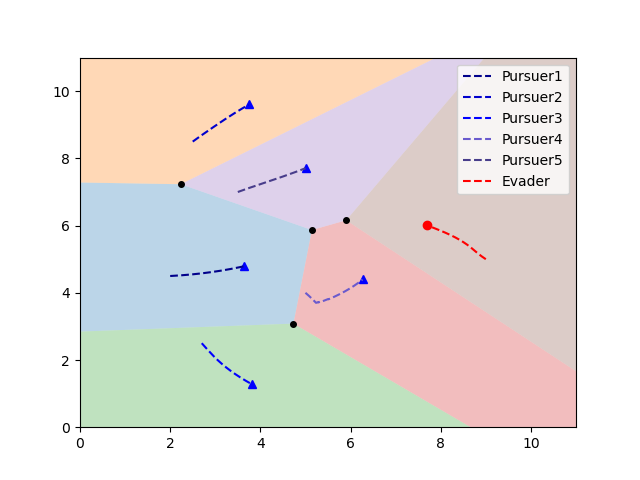}
         \caption{$t=1.67$s}
         \label{fig:3x1_dc_3}
     \end{subfigure}
     \hfill
     \begin{subfigure}[b]{0.3\textwidth}
         \centering
         \includegraphics[width=\textwidth]{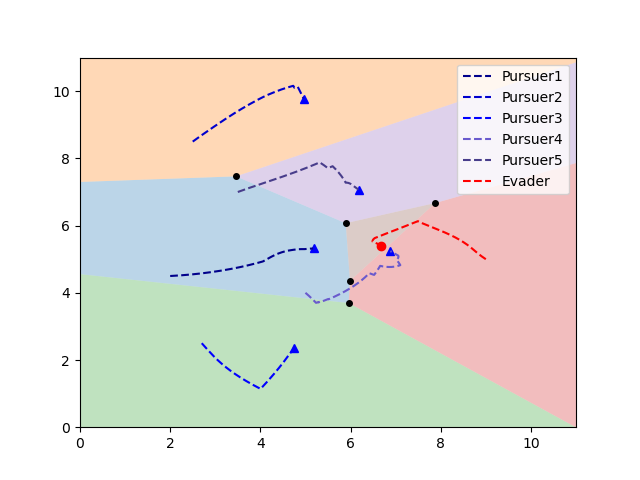}
         \caption{$t=3.34$s}
         \label{fig:5x1_dc_3}
     \end{subfigure}
        \caption{Trajectories for 5 pursuers(blue triangle) and 1 evader (red circle) under the area minimization strategy. Each shaded area represents Voronoi cell of the related point. }
        \label{fig:all_5x1_voronoi}
\end{figure}

Fig. \ref{fig:all_1x1_voronoi}, \ref{fig:all_3x1_voronoi}, and \ref{fig:all_5x1_voronoi} demonstrate the simulation results for three different time steps of which the last figures illustrate the capture moment. Even though capture condition is satisfied as shown in Fig. \ref{fig:all_1x1_voronoi}, evader acts unintelligently due to constant area policy. In Fig. \ref{fig:all_3x1_voronoi} and Fig. \ref{fig:all_5x1_voronoi}, more complex situation is handled; therefore, as mentioned in the Section 2, the strategy of evader applied in the first case is switched to move-to-centroid policy. 
In order to observe the correlation between the number of pursuers and capture time, 50 Monte-Carlo simulations are performed and corresponding capture times are gathered as represented in the Table. \ref{table:an_tima_capt}. 
It is shown that the more pursuers are involved in the game, the less time is required for capture.

\begin{comment}
1-1 \\ pursuer moves toward the centroid of Voronoi boundary
evader: area maximization: to keep the safe reach area constant, moves toward the centroid of Voronoi boundary evader \textit{show unintelligent behaviour even though capture condition is met in all cases}
3-1\\5-1 pursuer: through voronoi tessellation/decomposition, evader moves the centroid of its voronoi cell/area 
5-1\\ number of pursuers 

\textbf{monte -carlo}\\
50 monte simulations are run for each case 
as the number of pursuers increases avg capture time decreases \end{comment}

\begin{comment}
In these simulations, pursuers try to minimize safe reachable area of the evader and move towards to centroid of the shared Voronoi boundary using \ref{eq:8}. \\
\end{comment}

\begin{table}[H]
\caption{Capture time comparison results obtained by MC simulations}
\begin{center}
\begin{tabular}{|c|c|}
    \hline
    \multicolumn{2}{c}{Area Minimization Strategy}\\
    \hline
    Case & Average time(sec)\\
    \hline
    1 pursuer & 3.3103\\
    \hline
    3 pursuers & 3.276\\
    \hline
    5 pursuers & 3.222\\
\hline
\end{tabular}
\end{center}
\label{table:an_tima_capt}
\end{table}
For each MC simulations, the initial conditions of each pursuer is drawn from a uniform distribution $x\in [3,5]$ and  $y\in [0,10]$; and for the evaders $x\in [7,9]$ and  $y\in [0,10]$.The tabulated results are shown in Table\ref{table:an_tima_capt}.

% \begin{algorithm}
% \caption{Global Single Evader Policy} \label{alg:TD3} 
% \While{$(\min _{j \in P_k} \Vert x_p^j - x_e^\kappa \Vert > r_c)$}{
% Calculate Voronoi tessellation with all agents \\
% Determine pursuers which are the Voronoi neighbours, \mathcal {N}_e \\
% \uIf{$x_p^j \in \mathcal {N}_e$}{
%     Compute control input \ref{eq:u_area_min} \;
%   }
%   \Else{
%     Compute control input \ref{eq:10} \;
%   }}
% \ENDFOR{}
% \ENDIF{}
% \ENDELSE{}
% \end{algorithm}

\subsection{High-Level Pursuit Evasion Allocation with Reinforcement Learning}

The simulations are conducted for $N=3$, on $3x3$ grid map. Defender's distribution (the controlled swarm) is concentrated at a base location at the left-center grid of the map. The intruders (adversarial swarm) are distributed to the right side of the map and their location is randomly selected. After first simulation step, the intruders move to the left to reach left side of the map. The main goal is to see if low level pursuit evasion games and their result change high level allocation strategies.

We employ Advantage Twin Delayed Deep Deterministic Policy Gradient (TD3) algorithm as shown in Algorithm \ref{alg:TD3} due to its stability on continuous action environments. TD3 is off policy algorithm that improves  DDPG algorithm with three major improvement: clipped double Q-Learning, delayed policy update and target policy smoothing. \cite{fujimoto2018addressing} \cite{TD3} For training agents, we use Stable Baseline3 \cite{TD3} and fully connected neural network with 2 layers. The layers have 400 and 300 neurons on both actor network and critic networks.

\begin{figure}[H]
    \centering
    \includegraphics[width=\textwidth]{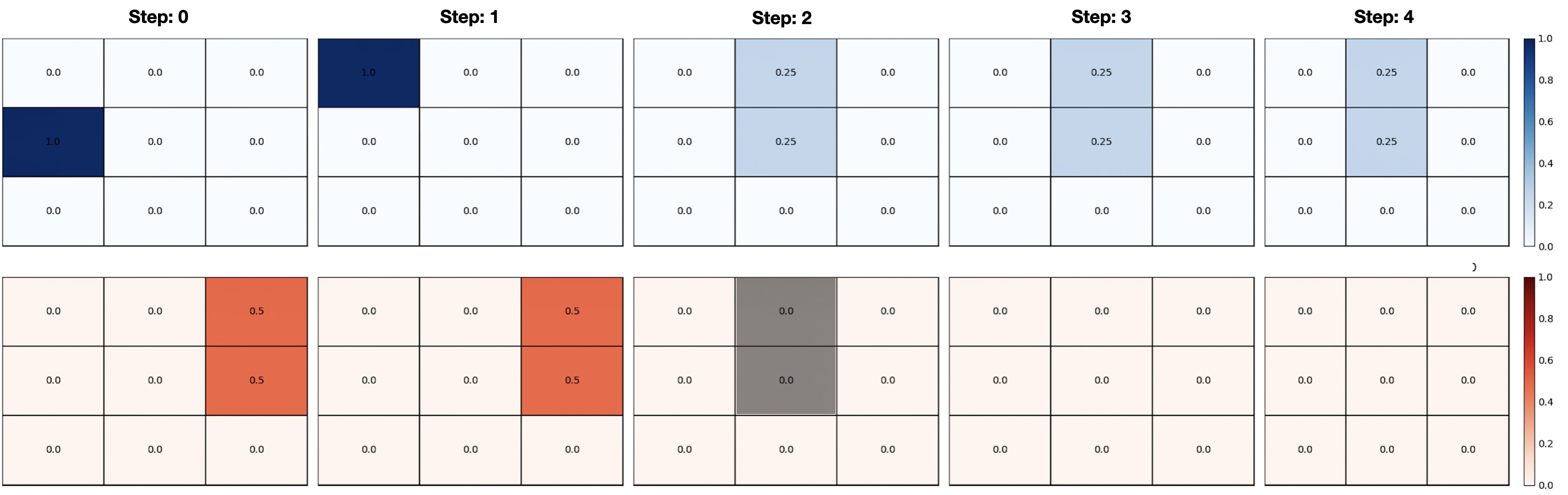}
    \caption{Density distributions throughout an episode: 100 defenders(blue), 50 intruders(red), Reward Coefficients: $C_{distribution} = 1$, $C_{capture} = 0$ }
    \label{fig:rl_result_1}
\end{figure}

In the Fig. \ref{fig:rl_result_1}, we only use distribution reward to see how agents behaves in the absence of pursuit evasion game information. The agent only collects reward as long as the defender distribution overlaps with the intruder distribution. On the $step=2$ of Fig. \ref{fig:rl_result_1}, it can be seen that the defender distributes its distribution evenly to destroy intruders in the time step without considering any strategic advantage. The results are intuitive, since only goal is to eliminate the intruder distribution as soon as possible.

\begin{figure}[H]
    \centering
    \includegraphics[width=\textwidth]{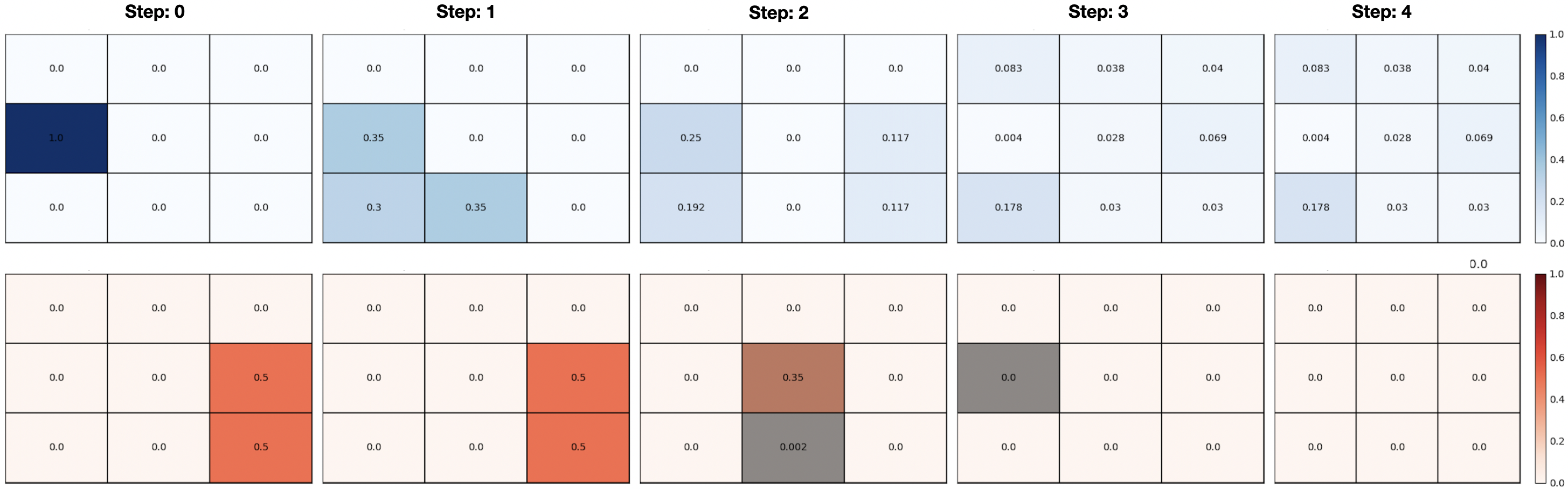}
    \caption{Density distributions throughout an episode: 100 defenders(blue), 50 intruders(red), Reward Coefficients: $C_{distribution} = 1$, $C_{capture}=1.5$}
    \label{fig:rl_result_2}
\end{figure}

The results of pursuit evasion games that stated at Section \ref{sec:pusuitresults} show that engaging with evaders with different number of pursuers significantly change overall outcome. In the Figure \ref{fig:rl_result_2}, it can be seen that, in the presence of the capture reward, defender strategy changes to destroying intruder in a sequential manner to suppress number of intruders in an engagement grid.

\begin{figure}[H]
     \centering
     \hfill
     \begin{subfigure}[b]{0.495\textwidth}
         \centering
    \includegraphics[width=\textwidth]{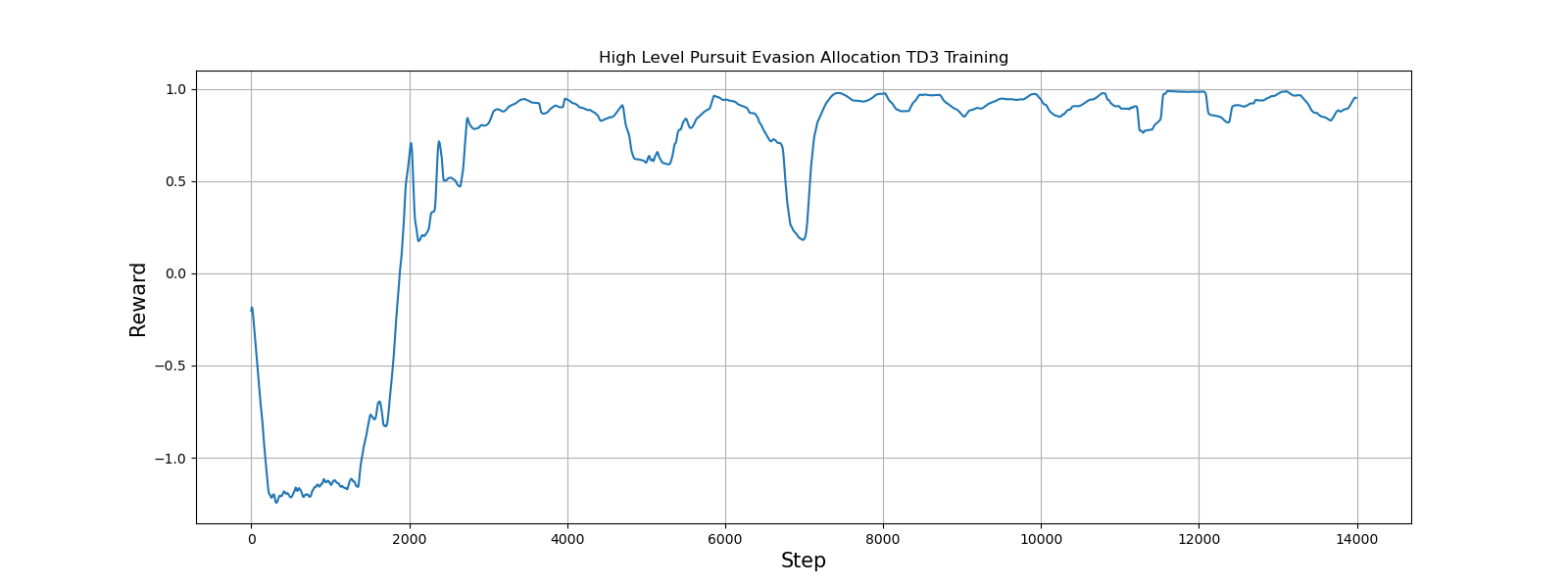}
    \caption{TD3 Training for High Level Allocation Map, Reward Coefficients: $C_{distribution} = 1$, $C_{capture}=0$, Maximum environment reward is 1 }
         \label{fig:rl_result_training_1}
     \end{subfigure}
     \hfill
     \begin{subfigure}[b]{0.495\textwidth}
    \centering
    \includegraphics[width=\textwidth]{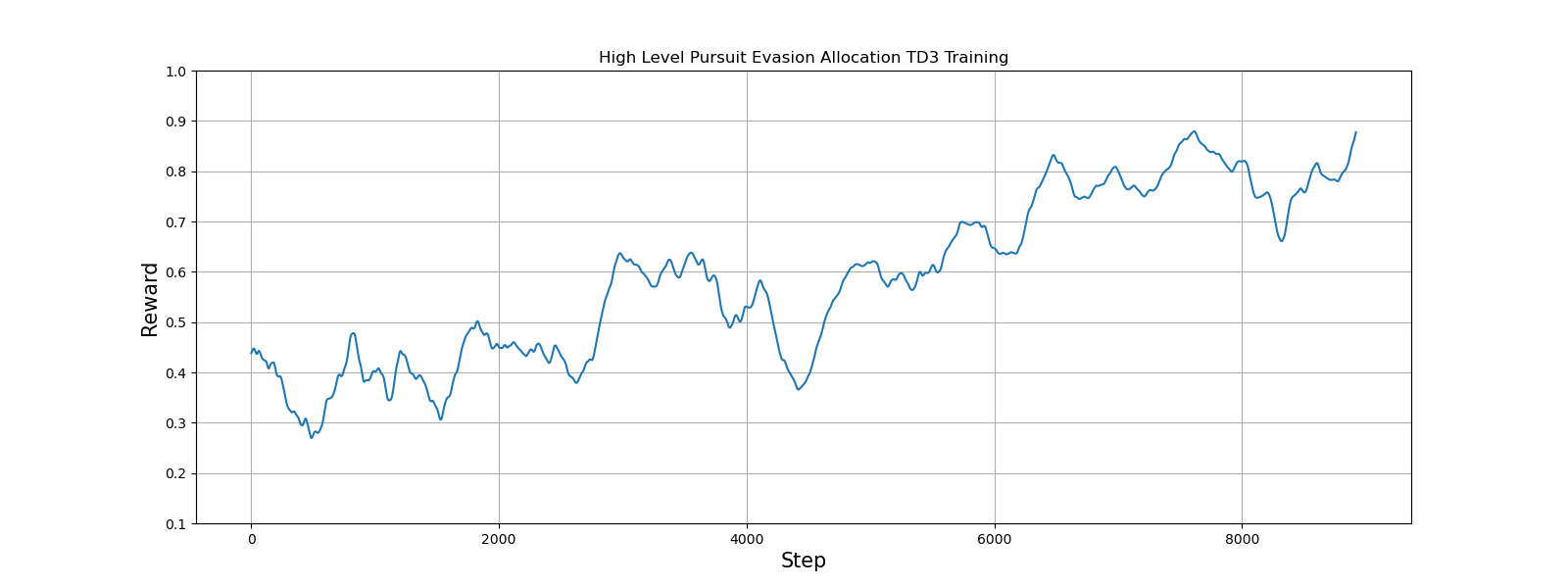}
    \caption{TD3 Training for High Level Allocation Map, Reward Coefficients: $C_{distribution} = 1$, $C_{capture}=1.5$, Maximum environment reward is 1 }
         \label{fig:rl_result_training_2}
     \end{subfigure}
     \hfill
    \label{fig:rldymmt}
\end{figure}
As can be seen at Figure \ref{fig:rl_result_training_1} and Figure \ref{fig:rl_result_training_2}, rewards of TD3 algorithm converges to maximum possible value in an episode even after several minutes of training.

\begin{figure}[H]
    \centering
    \includegraphics[width=\textwidth]{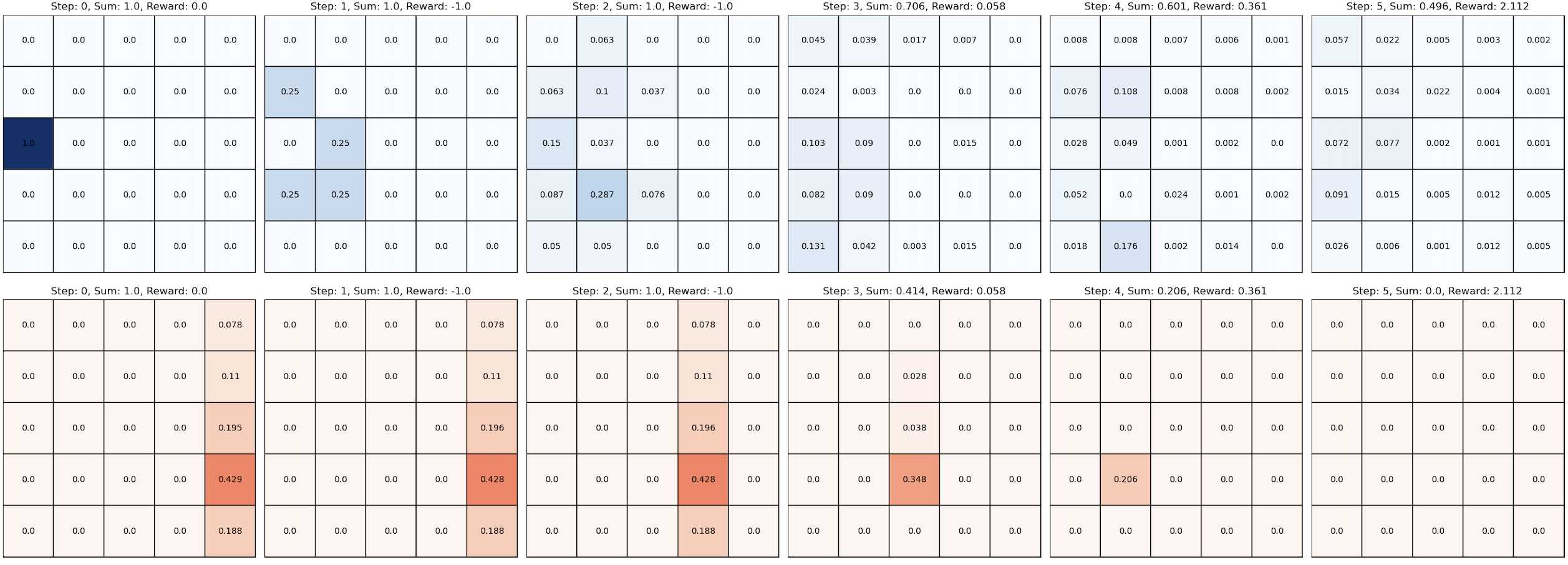}
    \caption{Density distributions throughout an episode: 100 defenders(blue), 100 intruders(red), Reward Coefficients: $C_{distribution} = 0.3$, $C_{capture}=1$}
    \label{fig:rl_result_3}
\end{figure}

\begin{figure}[H]
    \centering
    \includegraphics[width=\textwidth]{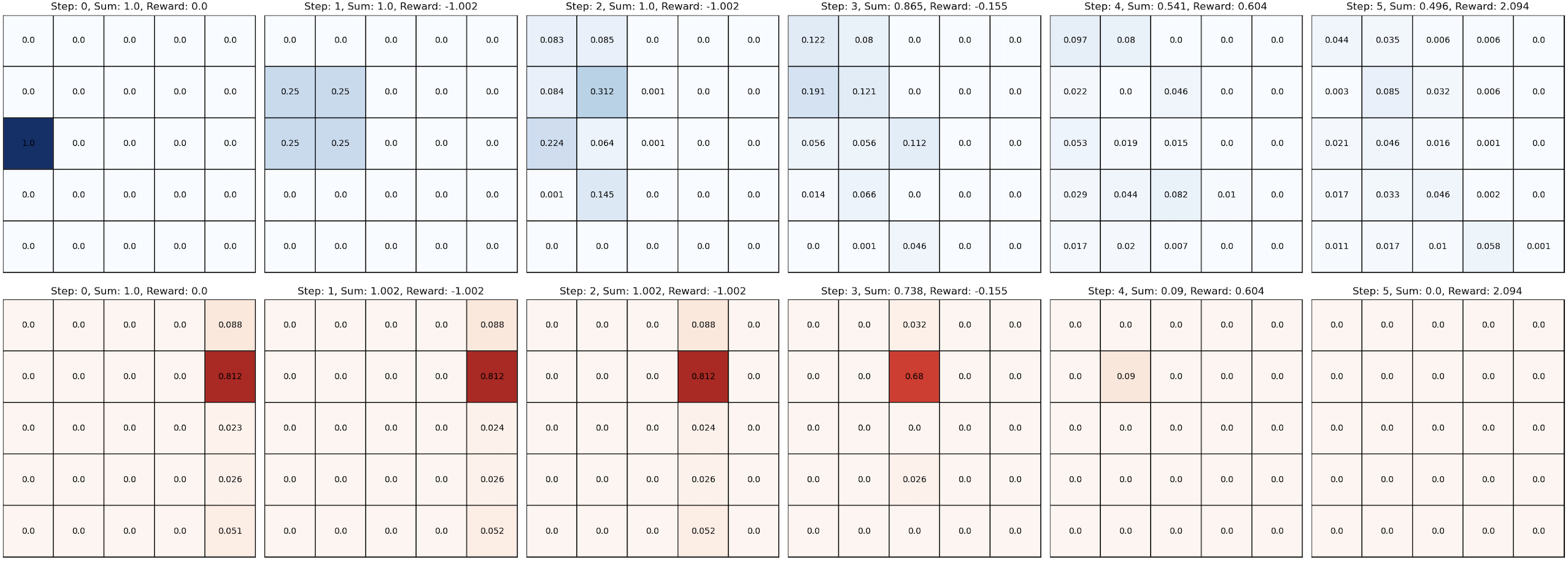}
    \caption{Density distributions throughout an episode: 100 defenders(blue), 100 intruders(red), Reward Coefficients: $C_{distribution} = 0.3$, $C_{capture}=1$}
    \label{fig:rl_result_4}
\end{figure}

The work can easily be extended to larger maps as seen at Figures \ref{fig:rl_result_3} \ref{fig:rl_result_4}. Since normalized distributions are used on the map, a solution can be found by increasing the number of grid on the map if desired. Another approach is that while keeping map size same, it can be solved as dividing to sub regions and solving every region separately. 
\section{Conclusion}
In this work, we have developed a framework that trains an RL agent to allocate defender swarm to intruder swarm. A grid map environment is created to match the swarm density distributions in small grids. Then, the problem is reduced to multiple pursuit-evasion games and corresponding rewards are gathered from different engagement scenarios on each grid at the lower level. It has been observed that the RL agent has learned to send the swarms to the appropriate grids for the scenarios in which they will be more successful in simulation results. As future work, we will develop strategies for large scale grid maps with different pursuit-evasion tactics to achieve state-of-the-art scalable swarm engagement framework.

% \begin{table}[H]
% \centering
% \begin{tabular}{|c| c|}
% \hline
% & \multicolumn{2}{ c}{Pure Pursuit Strategy} \\ 
% \hline
% Case & Time capture \\
% \hline
% 1x1 & 0.037\\
% \hline
% 3x1 & 0.037\\
% \hline
% 5x1 & 0.037\\
% \hline

% \end{tabular}
% \caption{\label{tab:table10} Capture time comparison for pure pursuit strategy}
% \end{table}

\bibliography{sample}

\end{document}